\documentclass[runningheads]{llncs}
\usepackage{graphicx}

\usepackage{soul}
\usepackage{xcolor}

\usepackage{support-caption}
\usepackage{caption}
\usepackage{subcaption}

\begin{document}

\title{Beyond Random Split for Assessing Statistical Model Performance}
\author{Carlos Catania \and
Jorge Guerra \and
Juan Manuel Romero \and
Gabriel Caffaratti and Martin Marchetta}

\institute{Universidad Nacional de Cuyo. \\Facultad de Ingeniería. LABSIN. \\Mendoza. Argentina}

\maketitle              
\pagestyle{plain}
\begin{abstract}
Even though a train/test split of the dataset randomly performed is a common practice, could not always be the best approach for estimating performance generalization under some scenarios. The fact is that the usual machine learning methodology can sometimes overestimate the generalization error when a dataset is not representative or when rare and elusive examples are a fundamental aspect of the detection problem. In the present work, we analyze strategies based on the predictors' variability to split in training and testing sets. Such strategies aim at guaranteeing the inclusion of rare or unusual examples with a minimal loss of the population's representativeness and provide a more accurate estimation about the generalization error when the dataset is not representative. Two baseline classifiers based on decision trees were used for testing the four splitting strategies considered. Both classifiers were applied on CTU19 a low-representative dataset for a  network security detection problem. Preliminary results showed the importance of applying the three alternative strategies to the Monte Carlo splitting strategy in order to get a  more accurate error estimation on different but feasible scenarios.

\end{abstract}

\section{Motivation}
An experimental design is a fundamental part of the machine learning workflow.  In the particular case of prediction problems, part of the design includes estimating the model's generalization performance. Estimating this performance is a critical aspect of developing a  model since it gives an idea of whether it can deal with future (not seen) scenarios reasonably well.

\begin{figure}[ht]
\centering
\includegraphics[width=0.75\textwidth]{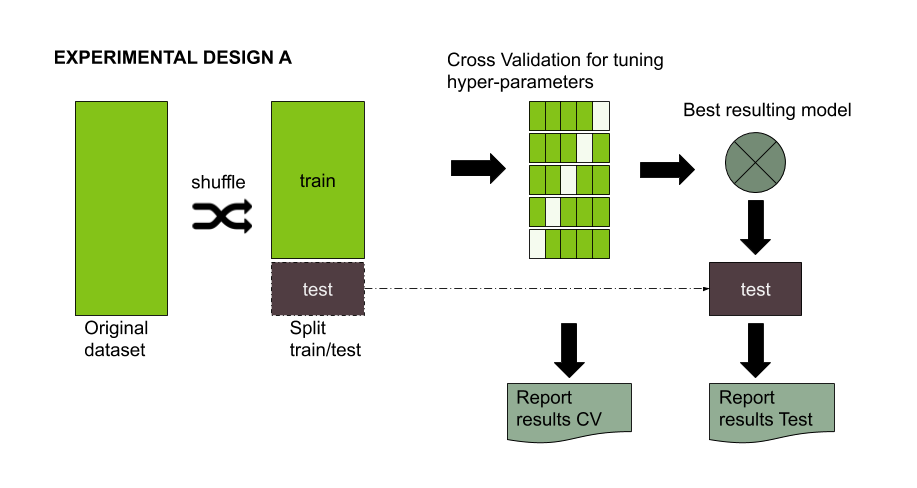}
\caption{ Standard experimental design for evaluating the performance of a machine learning model.}
\label{fig:ml-experimental-designA}
\end{figure}

The standard experimental design for evaluating the performance of a machine learning model is well known. As depicted in  Figure~\ref{fig:ml-experimental-designA} a dataset is split usually in a 70/30 ratio. 30\% of the data, called testing set, should be left aside and ideally never used until model tuning has finished. On the other hand, 70\% of the data, referred as training set, could be used to train and optionally validate the model or conduct a hyperparameter search for model tuning.

Train and test datasets need some degree of similarity (both need to follow the same distribution). Otherwise, it would be impossible for the model to achieve a decent performance on  test. However, if the examples are too similar in both datasets, then it is not possible to assure an accurate generalization performance for the model. Moreover, the model overall generalization performance could be overestimated. 

Even though a train/test split of the dataset randomly performed is a common practice, is not always the best  approach for estimating performance generalization under some scenarios. A common situation is when predicting patient outcomes. In these cases, the model should be constructed using certain patient sets (e.g., from the same clinical site or disease stage), but then need to be tested on a different sample population \cite{Kuhn2013}. Another situation is the fact that it is not always possible to have access to a representative sample. Detecting a non-representative sample is possible through the application several techniques, such as cross-validation, learning curves, and confidence intervals, among others. Unfortunately, in many cases, a non-representative sample is all we have to generate a machine learning model. In those cases  when a sample does not follow the same population distribution, a random split might not provide the required level of representativeness for rare or elusive examples in a testing set. As a result, the standard error metrics could overestimate the performance of the model. In classification problems, it is possible to deal with the lack of representativeness using a stratification strategy. However, when rare examples are not labeled, a predictor-based sampling strategy will be necessary \cite{Willett1999,Clark1997}

In the present work, we analyze several strategies based on the predictors' variability to split in training and testing sets. Such strategies aim at guaranteeing the inclusion of rare or unusual examples with a minimal loss of the population's representativeness. The hypothesis is that by including rare examples during model evaluation a more accurate performance estimates will be obtained.

The contributions of the present article are:
\begin{itemize}
    \item The analysis of four splitting strategies with different distributions for training and testing sets.
    \item The evaluation of two different tree-based baseline classifiers over four different splitting strategies. 
\end{itemize}


\section{Splitting Strategies}
\label{sec:sampling-strategies}

\subsection{Monte Carlo}

The usual strategy for model evaluation consists of taking an uniformly random sample without replacement of a portion of the data for the training set, while all other data points are added to the testing set. Such strategy can be thought as special case of the Monte Carlos Cross Validation (MCCV) \cite{xu2001} with just one resample instance. The Monte Carlo (MC) splitting strategy guarantees the same distribution across not only response but also predictor variables for training and testing sets. 
 
In comparison with Monte Carlo, the remaining splitting strategies provide steps to create different test sets that include rare or elusive examples while maintaining similar properties across the predictor space as the training set.

\subsection{Dissimilarity-based}
Maximum dissimilarity splitting strategies were proposed by \cite{Clark1997} and \cite{Willett1999}. The simplest method to measure dissimilarity consist of using the distance between the predictor values for two samples. The larger the distance between points, the larger indicative of dissimilarity.  The application of dissimilarity during data splitting requires a set initialized with a few samples. Then, the dissimilarity between this set and the rest of the unallocated samples can be calculated. The unallocated sample that is most dissimilar to the initial set would then be added to the test set.

Dissimilarity splitting proved to be useful over chemical databases' splitting \cite{Snarey1997,Martin2012}. Nevertheless, this method strongly depends on the initial set used to calculate the dissimilarity of the rest of the samples, prompting problems in cases of small datasets where the initial set is not representative enough \cite{Yang2019}.

\subsection{Informed-based}
A well-known non-random split strategy consists of using some kind of grouping information from the data to restrict the set of samples used for testing. The general idea after splitting the data is that members of a group present in training set should not be included in the testing set. Such strategies are well-known in areas such as Medicine and Finances \cite{Kuhn2013}, where testing should be conducted on a different patient group or, in the finances field, where the model should be tested on a time series from a different time period.

\subsection{Clustering-based}
The clustering split strategy follows the same principle of the informed split. However, there could be situations where no grouping information can be extracted from the samples to perform an informed split. In these cases, the application of a clustering algorithm could be used to replace the missing information. The  labels generated by this procedure will be then used for performing a group split similarly to the informed split strategy.

\section{Application to a Network Security Dataset}
The four splitting strategies described in section~\ref{sec:sampling-strategies} were applied to a network security dataset for botnet detection conformed by nineteen network captures published  by the stratosphere IPS research group at CTU \cite{CTU19}. Specifically, the dataset has fourteen botnet captures and five normal captures  including traffics like DNS, HTTPS and P2P.  In total, all captures represents 20866 connections having 19271  labeled as ``Botnet" and 1595 labeled as ``Normal". All these captures were gathered between 2013 and 2017.


The first ten predictors of the CTU19 dataset summarize each flow of connections with same IP, protocol and destination port into a 10-dimensional numerical vector: 

\begin{equation}
\label{eq:data_feat_vec_eq}
\mathbf{v}_{feat} = <x_{sp}, x_{wp}, x_{wnp}, x_{snp}, x_{ds}, x_{dm}, x_{dl}, x_{ss}, x_{sm}, x_{sl}>
\end{equation}

\noindent wherein the first four dimensions of the numerical vector represent the periodicity predictors (strong periodicity ($x_{sp}$), weak periodicity ($x_{wp}$), weak non periodicity ($x_{wnp}$) and strong non periodicity ($x_{snp}$)), the  next three refer to the duration predictors (duration short ($x_{ds}$), duration medium ($x_{dm}$) and duration large ($x_{dl}$), respectively), and the last three represent the size predictors (size short ($x_{ss}$), size medium ($x_{sm}$), size large ($x_{sl}$)). The vector for a given connection is generated considering the cumulative frequency of the corresponding values associated with the behavioral of each predictor.




In addition to the information provided by the flow-based predictors, the CTU19 dataset includes information related with the flow like source IP, destination IP, protocol, port and the source linked with each capture. However, the present study will focus only on the information provided by the flow-based predictors as discussed in \cite{guerra2019,guerra2019b}.

\subsection{Initial Exploratory Analysis}
\begin{figure}[!ht]
\centering
\includegraphics[width=1\textwidth]{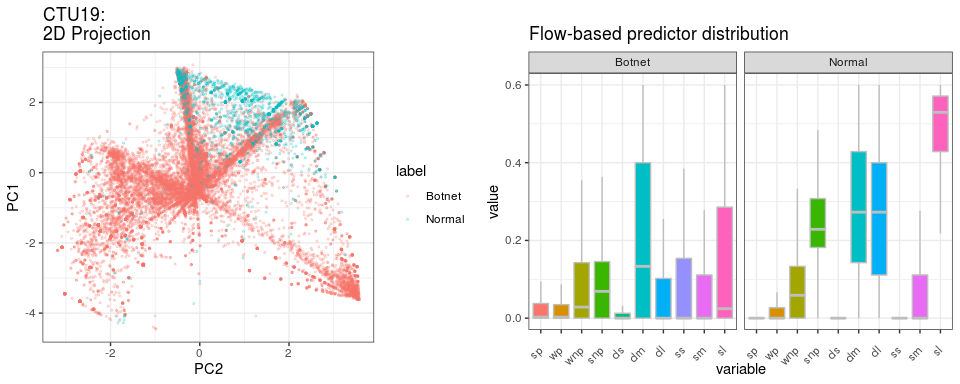}
\caption{ Boxplot and 2D projection of the flow-based predictors in the CTU19 dataset. Botnet traffic in red and Normal in blue.}
\label{fig:ctu19projection}
\end{figure}

Fig.~\ref{fig:ctu19projection} represents a 2D projection of the CTU19 dataset considering the first 2 principal components (PCA was applied to the flow-based predictors). In addition, the figure includes the distribution for each predictor. As depicted by the box plot, botnet and normal classes present a different distribution over the flow-based predictors. However, both classes show a partial overlap at the time of projecting them into a 2D space.

\begin{figure}[!ht]
\centering
\includegraphics[width=1\textwidth]{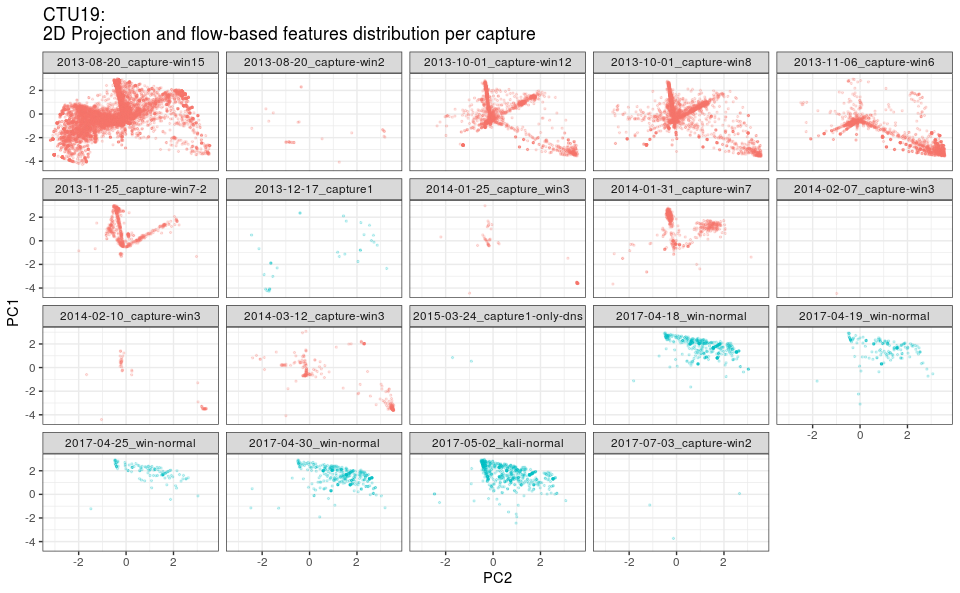}
\includegraphics[width=1\textwidth]{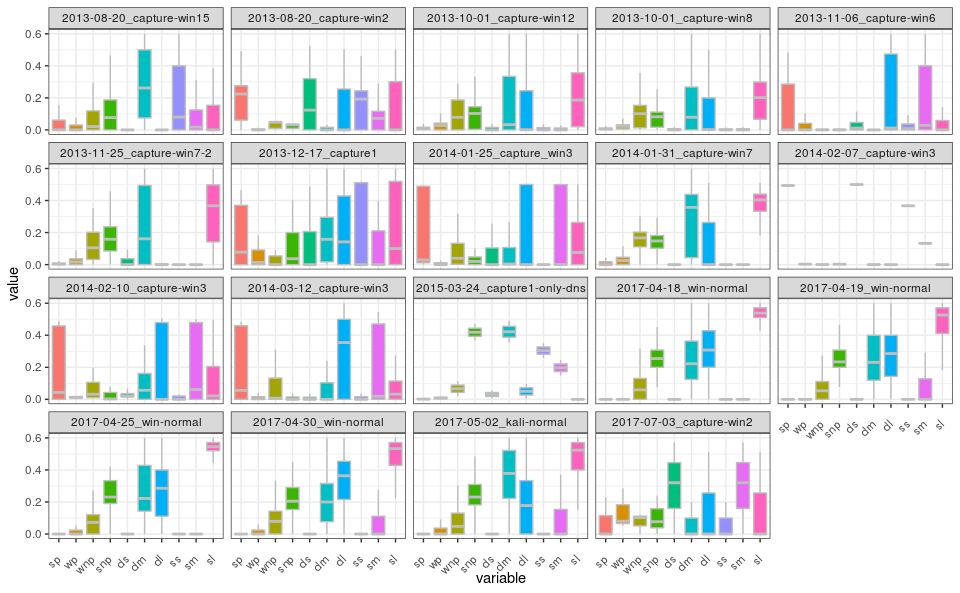}
\caption{On the top, the scatter plots on the top depicts the 2D projection using PCA for each one of the 19 captures conforming the CTU19 dataset (Botnet traffic in red and Normal in blue). On the bottom, the corresponding distributions for the 10 flow-based predictors on each capture.}
\label{fig:ctu19projection_bydataset}
\end{figure}

\noindent Fig.~\ref{fig:ctu19projection_bydataset} decomposes the CTU19 dataset by traffic capture with the same information as provided by Fig~\ref{fig:ctu19projection}. In general, the patterns from the 2D projection normal captures are  different compared with Botnet captures. Normal captures are concentrated while Botnet captures spread along the 2D predictor space. When analysing Normal captures, most of them overlap the same predictor space. In other words, each capture has some examples on every portion of the Normal predictor subspace, which suggests the adequate representativeness. Nevertheless, in the case of Botnet, there are several cases of captures having only a limited presence on the class predictor subspace (see captures 2014-02-07-win3 and 2014-01-25-win3). Such lack of representativeness observed by some Botnet captures could difficult the classification model's performance estimation.


\begin{figure}[!th]
\centering
\includegraphics[width=1\textwidth]{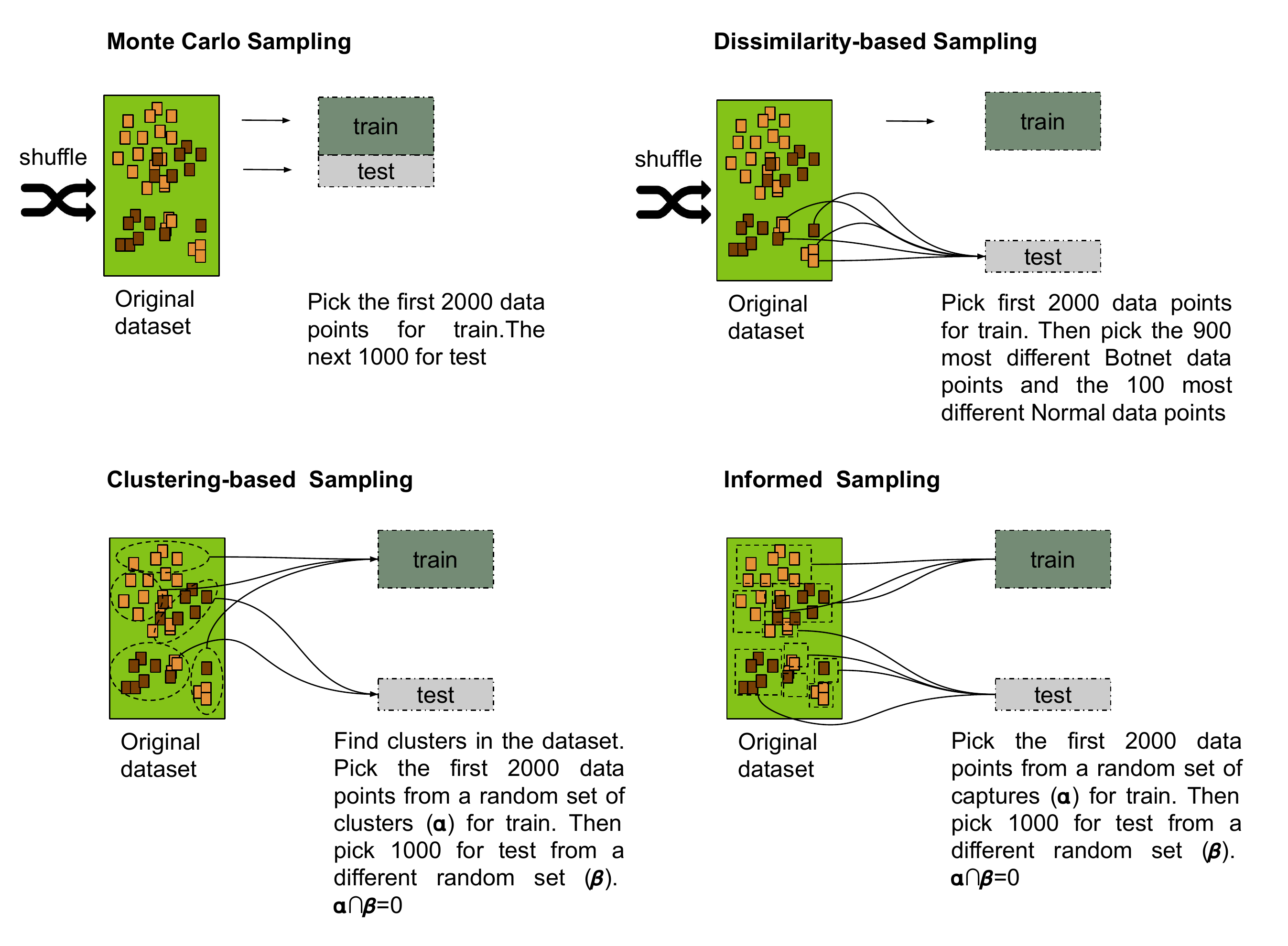}
\caption{Procedure used for generating the training and testing sets for the four splitting strategies: Monte Carlo, Dissimilarity-based, clustering-based and informed-based. In all the cases a training set with 2000 data points and a testing set with 1000 were generated. }
\label{fig:sampling-strategies}
\end{figure}

\subsection{Training and Testing Sets Creation}
Fig. \ref{fig:sampling-strategies} describes the process used for the creation of the training and testing sets according to each splitting strategy. A subset of 3000 data points from CTU19 were used for each strategy. 2000 data points for conforming the training set and 1000 for the testing set. The generation of the training and testing sets will depend on the splitting strategy applied as discussed in section \ref{sec:sampling-strategies}. The procedure is repeated 25 times. Therefore 25 pairs of training and testing sets are generated for each splitting strategy. 

\begin{figure}[ht]
\centering
\includegraphics[width=1\textwidth]{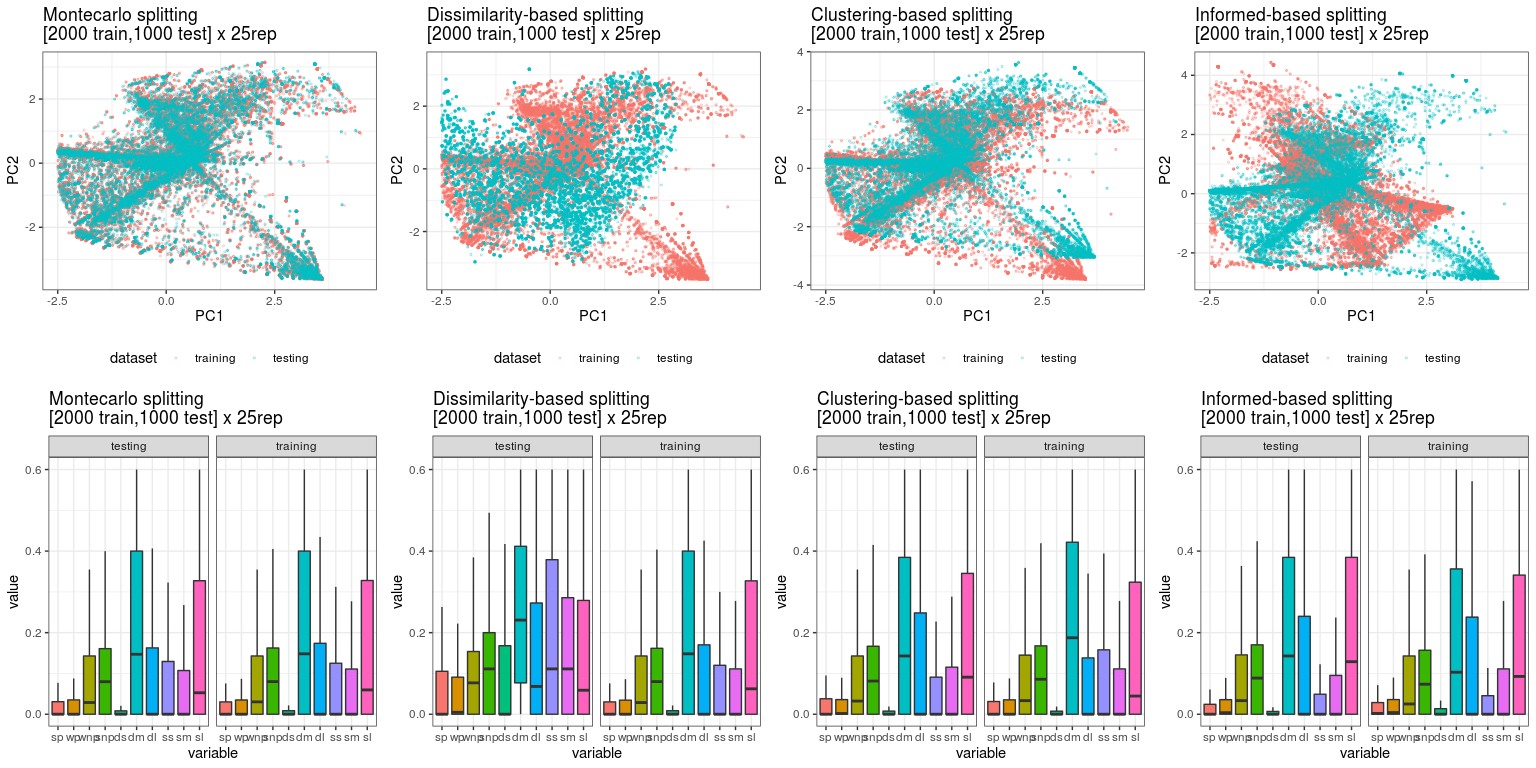}
\caption{ 2D projection and boxplot distribution for the 25 pairs of training and testing datasets for each splitting strategy.}
\label{fig:sampling-projection}
\end{figure}

\noindent The differences between each splitting strategy are observed in Fig~\ref{fig:sampling-projection} considering data points from the 25 samples. The 2D projection using the first two principal components confirms the similarities between training and testing sets when the baseline Monte Carlo splitting is applied. Moreover, both datasets follow the same pattern in the predictor space which corresponds with the similarity observed in the predictor distributions (see box plots below). However, different patterns are observed between training and testing sets in the remaining splitting strategies. In particular,  dissimilarity-based and informed-based splitting present the most different patterns. Nevertheless, the same pattern is observed in both datasets for the clustering-based splitting case, although with a small displacement from the axis. 

\noindent When projecting many predictors into two dimensions, intricate predictor relationships could mask regions within the predictor space where the model will inadequately predict new samples. 

An algorithmic approach  as described in \cite{hastie2008} is applied to get more detailed information about the similarities between training and testing datasets. The general idea is to create a new dataset randomly permuting the predictors from the training set and then, row-wise concatenate to the original training set labeling original and permuted samples. A classification model is run on the resulting dataset to predict the probability of new data being in the class of the training set. Table \ref{tab:train-prediction} shows average percentage of samples from testing set not recognized as part of the training set.

\begin{table}[ht]

\centering
\caption{Average Percentage and standard deviation of test samples not recognized as part of the training set for the four splitting strategies   }
\begin{tabular}{ccclll}
  \hline
 &\textbf{ Avg Err} \% & \textbf{sd }&& \textbf{Splitting Strategy} \\ 
  \hline
   & 0.06 & 0.05 && informed-based \\ 
   & 0.01 & 0.01 && monte carlo \\ 
   & 0.17 & 0.13 && cluster-based \\ 
   & 0.06 & 0.02 && dissimilarity-based \\ 
   \hline
\end{tabular}
\label{tab:train-prediction}
\end{table}

As expected, Monte Carlo splitting exhibits  the lower error and standard deviation, confirming the similarities between training and testing set observed in the 2D projection from Fig.\ref{fig:sampling-projection}.  Both dissimilarity and informed-based strategies has the same average error. However, informed-based shows a higher variation. Finally, the clustering-based strategy shows the biggest differences between training and testing sets.

\section{Error Estimation on Baseline Classifiers}
\label{sec:error-estimate-classifier}
The impact of the different splitting strategies discussed in section \ref{sec:sampling-strategies} is measured on Random Forest (RF) and CatBoost (CB) classifiers. Both CB and RF are two well-known classifiers  providing acceptable results on tabular data without conducting a hyper parameters tuning. Both baseline classifiers were executed with default parameters. Nevertheless, downsampling technique is applied to training set to deal with classes imbalance issues.

\subsection{Metrics}
Several standard metrics for model classification assessment were used to evaluate baseline classifiers performance on the different train/test datasets. The metrics correspond to True Positive Rate (TPR) and False Positive Rate (FPR). The \textbf{Sensitivity} measures the proportion of positives that are correctly identified (TPR), and the \textbf{Specificity} measures the proportion of negatives that are correctly identified ($1-FPR$).

Additional metrics were used to deal with class imbalance: \textbf{F1-Score} and \textbf{Balanced Accuracy}. F1-Score is computed as the weighted average between TPR and the total numbers of malicious connections in the dataset. Balanced Accuracy is calculated as the average of correctly classified proportion of each class individually.

\subsection{Results}

\begin{figure}[!ht]
\begin{subfigure}{.5\textwidth}
  \centering
  \includegraphics[width=.8\linewidth]{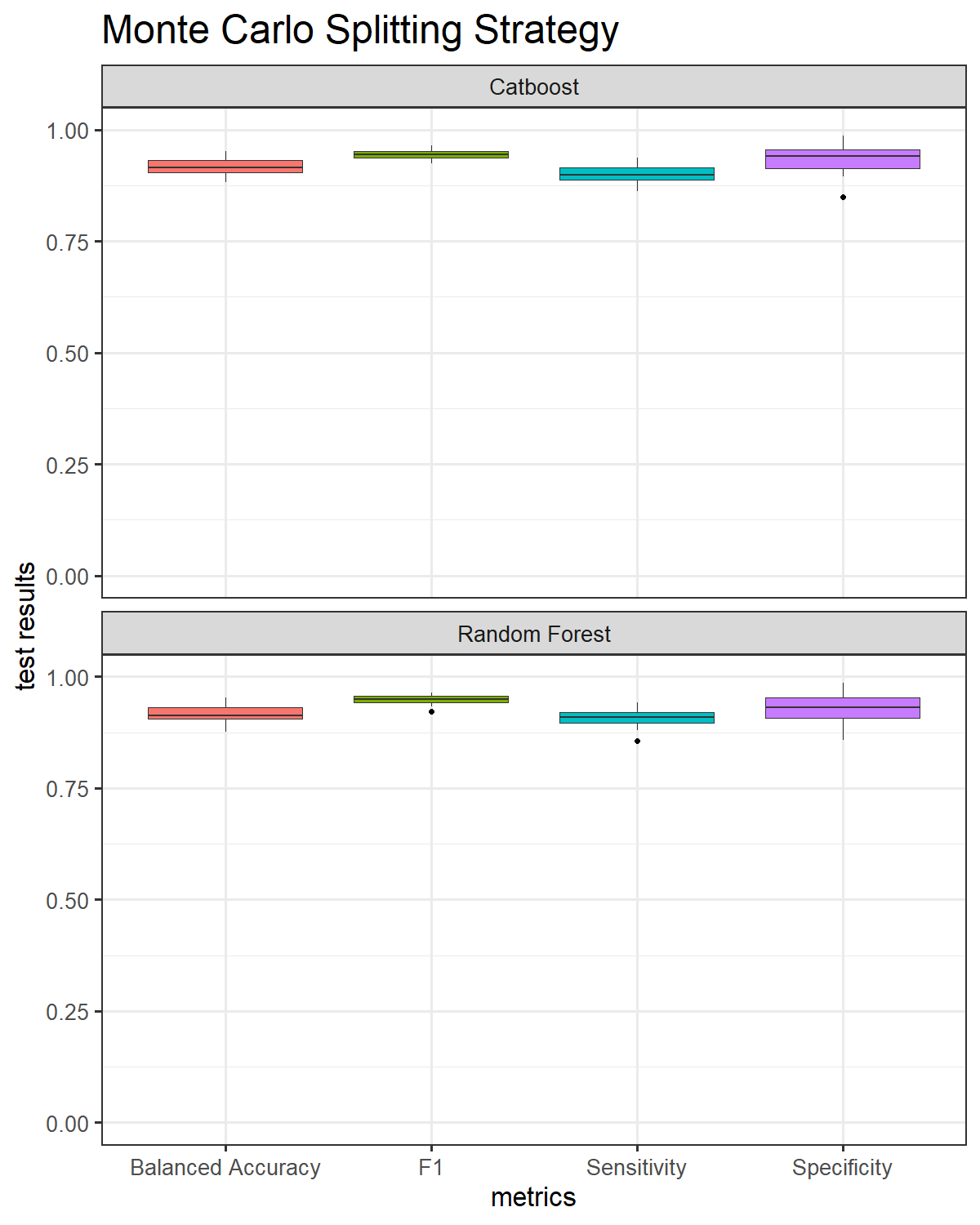} 
  \caption{Monte Carlo splitting Strategy}
  \label{fig:sub-first}
\end{subfigure}
\begin{subfigure}{.5\textwidth}
  \centering
  \includegraphics[width=.8\linewidth]{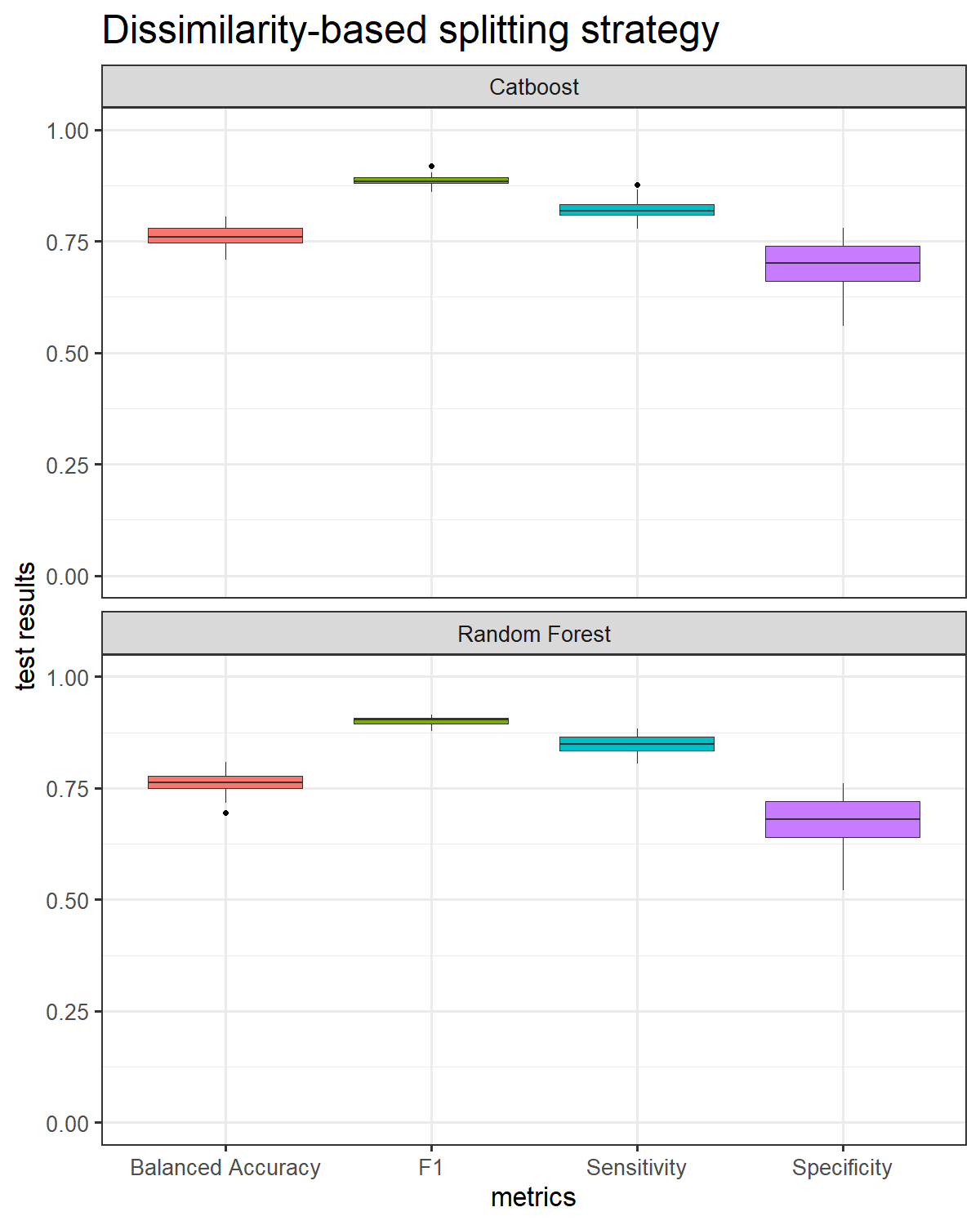}
  \caption{Dissimilar splitting Strategy}
  \label{fig:sub-second}
\end{subfigure}

\begin{subfigure}{.5\textwidth}
  \centering
  \includegraphics[width=.8\linewidth]{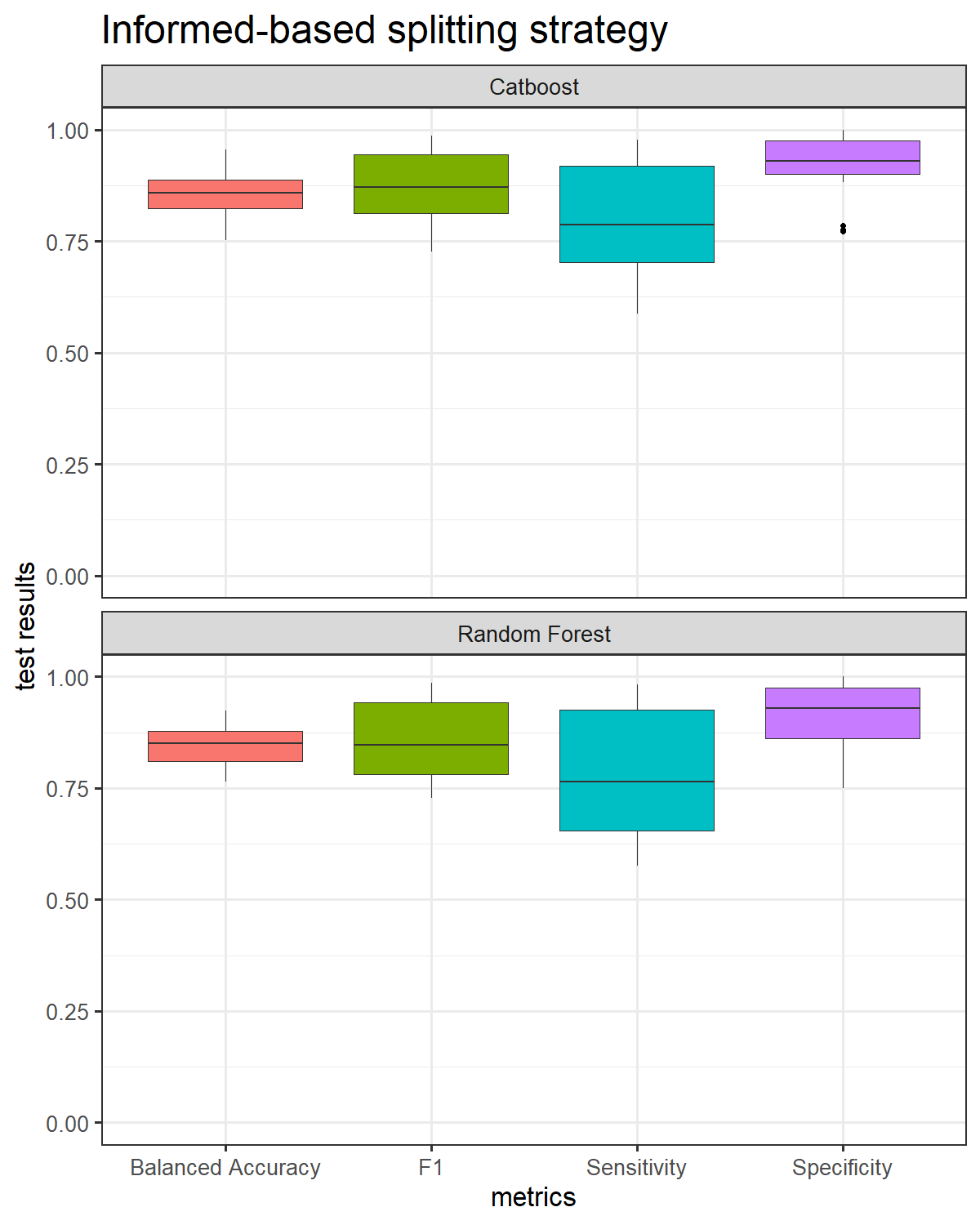}  
  \caption{Informed-based splitting Strategy}
  \label{fig:sub-third}
\end{subfigure}
\begin{subfigure}{.5\textwidth}
  \centering
  \includegraphics[width=.8\linewidth]{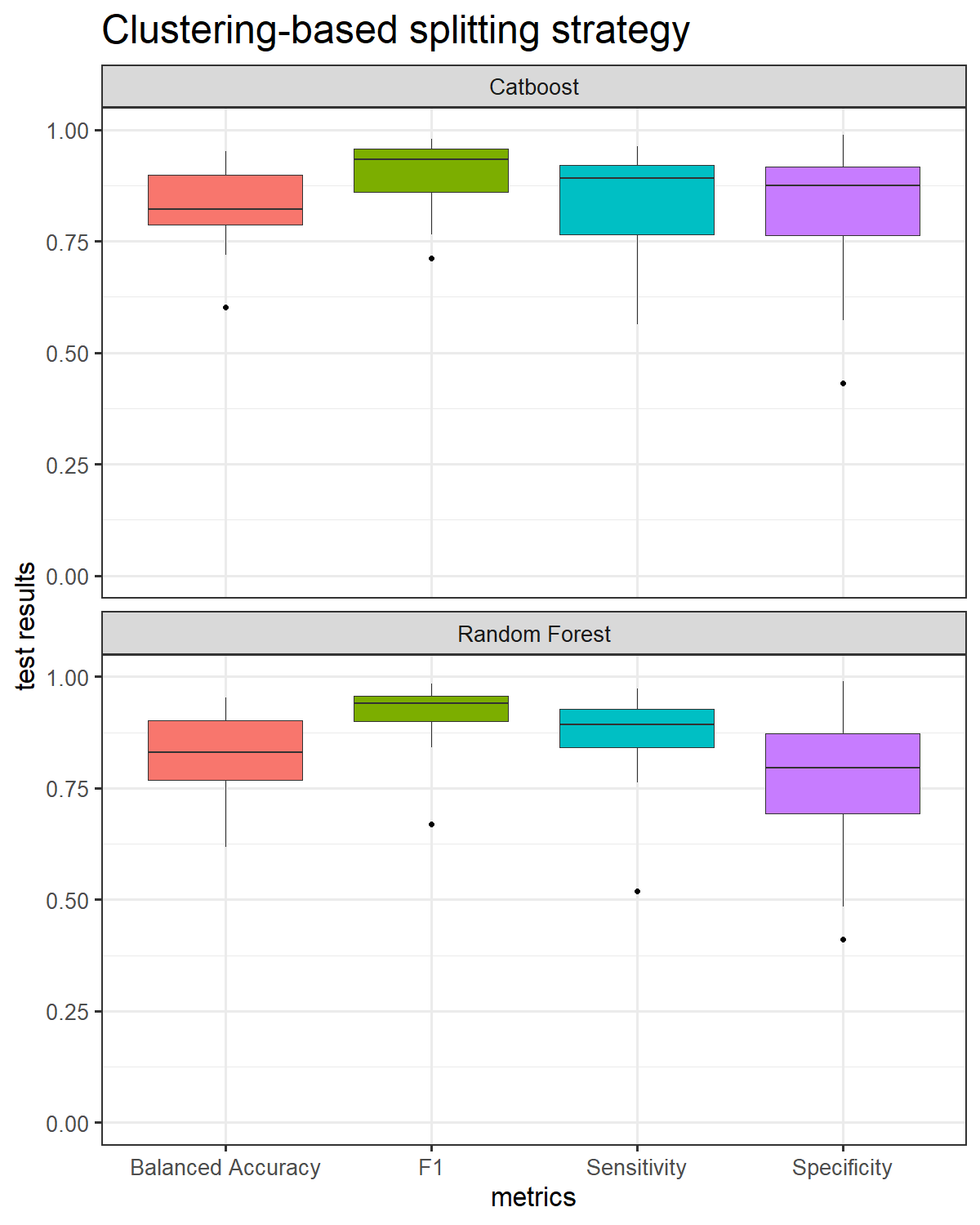}  
  \caption{Clustering-based splitting Strategy}
  \label{fig:sub-fourth}
\end{subfigure}
\caption{Random Forest and CatBoost models' performance result for each splitting strategy}
\label{fig:model_results}
\end{figure}

Fig.~\ref{fig:model_results} depicts the overall performance results for  baseline classifications models using the splitting strategies described in Section \ref{sec:sampling-strategies}. Both models were used to predict ``Botnet'' or `Normal'' connections. Since no significant difference is observable in the performance between both baseline models, the results discussed in this section corresponds only to the catboost algorithm. 

Fig.~\ref{fig:sub-first} displays testing set's results for RF (below) and CB (above) using the MC splitting strategy. Prediction performance in terms of Balanced Accuracy  median is around 0.91. The remaining metrics obtained similar values for both baseline models. A low variability is observed in all the considered metrics. For the Dissimilarity-based splitting strategy (see Fig.~\ref{fig:sub-second}), all considered metrics decrease in comparison to MC strategy. The Balanced Accuracy median decreases to 0.75 while the specificity decreases to 0.70. The F1 and Sensitivity median maintained higher values, although lower than the observed in the MC strategy. 

Fig.~\ref{fig:sub-third} states the results for the Informed-based splitting strategy. In this case, Balanced Accuracy and F1 show a median value of 0.85, 0.87, respectively. On the other hand, in terms of Specificity and Sensitivity, the median values are around 0.78 and 0.93, respectively. Notice that despite the good performance in terms of median values, a considerable variation is observed for the F1 and Sensitivity metrics. Finally, Fig.~\ref{fig:sub-fourth} presents the models' results using the Clustering splitting strategy. The Balanced Accuracy median value decreased to 0.82 while F1 increases to 0.93 compared with the informed-splitting strategy. The rest of the metrics, Sensitivity and Specificity, shown similar results with median around 0.89 and 0.87, respectively.

\section{Discussion}
\label{sec:discussion}
In general both baseline models presented acceptable performance for predicting ``Botnet'' and ``Normal'' classes since the median Balanced Accuracy metric was over 0.75 in most of cases. 
As expected, the Monte Carlo splitting strategy showed the smallest estimation error (a high balanced accuracy). The similarities between training and testing sets was already observed in Table \ref{tab:train-prediction}, where only 1\% of the testing set was not recognized as part of training set. When the size of not recognized samples increments only to a 6\%, such as in the case of dissimilarity strategies, the estimation error increases considerably (from 0.91 to 0.75 median Balanced Accuracy). Since Monte Carlo and Dissimilarity-based splitting strategies use the same procedure for generating the training set, their analysis can provide a valuable error estimation under an anomalous but certainly feasible set of samples.

In the case of informed and clustering based splitting strategies, the difference between training and testing sets is not only larger than Monte Carlo but also both show a larger variation (see Table~\ref{tab:train-prediction}). A considerable variation is also observed in the performance of both baseline classifiers. However, both splitting strategies still provide information about the robustness of the models on sets with different representativeness. Previous statement is particularly valid for the informed-based splitting strategy, where captures with very different representiveness levels are used for building training and testing sets. For instance, informed-based Balanced Accuracy range provides ad-hoc information about the expected values when not so representative sets are used for training. Moreover, it is possible to infer that a 0.75 value for Balanced Accuracy could be the worst performance scenario observed by the model. Such value is still suitable under some real-life situations. 

The clustering-based strategy provides a more extreme scenario than informed-based for estimating the model performance under not so representative sets. Under the clustering-base strategy a concentrated portion of the predictor space present in the testing set is excluded from training set, whereas in the informed-based it is possible to find datapoints spread along the whole predictor space. Consequently, it is possible to observe higher variation in the baseline models performance.

\section{Concluding Remarks and Future Work}
Despite being the standard splitting strategy, Monte Carlo can overestimate the results when dataset is not representative. Other splitting techniques based on dissimilarity, information present in the dataset and the application of a clustering algorithm can help in the estimation under different low-representativeness scenarios. 

Multiple training and testing sets where generated using the different strategies on the CTU19 Botnet dataset. Small differences between training and testing sets were corroborated using the algorithm proposed by \cite{hastie2008} in all the four techniques. As expected Monte Carlo showed the smallest differences while clustering-based showed the biggest.

Two baseline classifiers were used for evaluating each splitting strategy in the error estimation process. The Dissimilarity-based splitting strategy provided a valuable error estimation under an anomalous but certainly feasible set of samples. On the the other hand, informed-based strategy offers ad-hoc information about the expected values when sets not so representive are used for training, while clustering-based strategy emerges as an alternative to informed-based for estimating the model performance under low representativeness sets with more pessimist estimation.

Preliminary results showed the importance of applying the three alternative strategies to the Monte Carlo splitting strategy in order to get a more accurate error estimation on different but possible situations. However, given the particular low-representativeness nature of the botnet detection problem, a deeper analysis and evaluation on other datasets should be conducted.

\section{Acknowledgments}
The authors would like to thank the financial support received by SIIP-UNCuyo during this work. In particular the projects 06/B363 and 06/B374. In addition, we want to gratefully acknowledge the support of NVIDIA Corporation with the donation of the Titan V GPU used for this research.

\bibliographystyle{splncs04}
\bibliography{samplingstrats}
\end{document}